\documentclass[letterpaper, 10 pt, conference]{ieeeconf}
\IEEEoverridecommandlockouts                              

\usepackage{cite}
\usepackage{amsmath}
\usepackage{amssymb}
\usepackage{xcolor}
\usepackage{graphicx}
\usepackage{dblfloatfix}
\usepackage{multirow}
\usepackage{caption}
\usepackage{url}
\usepackage{lipsum}

\title{\Large \bf
Learning Invariant Representation of Tasks for Robust Surgical State Estimation}

\author{Yidan Qin$^{1,2}$, Max Allan$^{1}$, Yisong Yue$^{2}$, Joel W. Burdick$^{2}$, Mahdi Azizian$^{1}$
\thanks{$^{1}$Intuitive Surgical Inc., 1020 Kifer Road, Sunnyvale, CA,94086, USA}%
\thanks{$^{2}$Mechanical and Civil Engineering, Caltech, Pasadena, CA, 91125, USA}%
\thanks{Emails: Ida.Qin@intusurg.com, Mahdi.Azizian@intusurg.com}
}

\begin{document}

\maketitle
\thispagestyle{empty}
\pagestyle{empty}

\begin{abstract}
\noindent
Surgical state estimators in robot-assisted surgery (RAS) - especially those trained via learning techniques - rely heavily on datasets that capture surgeon actions in laboratory or real-world surgical tasks. Real-world RAS datasets are costly to acquire, are obtained from multiple surgeons who may use different surgical strategies, and are recorded under uncontrolled conditions in highly complex environments. The combination of high diversity and limited data calls for new learning methods that are robust and invariant to operating conditions and surgical techniques. We propose \emph{StiseNet}, a Surgical Task Invariance State Estimation Network with an invariance induction framework that minimizes the effects of variations in surgical technique and operating environments inherent to RAS datasets. StiseNet's adversarial architecture learns to separate nuisance factors from information needed for surgical state estimation. StiseNet is shown to outperform state-of-the-art state estimation methods on three datasets (including a new real-world RAS dataset: HERNIA-20).
\end{abstract}

\section{INTRODUCTION}

While the number of Robot-Assisted Surgeries (RAS) continues to increase, at present they are entirely based on teleoperation. Autonomy has the potential to improve surgical efficiency and to improve surgeon and patient comfort in RAS, and is increasingly investigated \cite{yang2017medical}. Autonomy can be applied to passive functionalities \cite{selvaggio2018passive}, situational awareness \cite{chalasani2018computational}, and surgical tasks \cite{shademan2016supervised, attanasio2020autonomous}. A key prerequisite for surgical automation is the accurate real-time estimation of the current surgical state. Surgical states are the basic elements of a surgical task, and are defined by the surgeon's actions and observations of environmental changes \cite{qin2020temporal}. Awareness of surgical states would find applications in surgical skill assessment \cite{tao2012sparse}, identification of critical surgical states, shared control, and workflow optimization \cite{padoy2019machine}. 

Short duration surgical states, with their inherently frequent state transitions, are challenging to recognize, especially in real-time. Many prior surgical state recognition efforts have employed only one type of operational data. Hidden Markov Models \cite{tao2012sparse, volkov2017machine}, Conditional Random Fields (CRF) \cite{tao2013surgical}, Temporal Convolutional Networks (TCN) \cite{lea2016temporal}, Long-Short Term Memory (LSTM) \cite{dipietro2016recognizing}, and others have been used to recognize surgical actions using robot kinematics data. Methods based on Convolutional Neural Networks (CNN), such as CNN-TCN \cite{lea2016temporal} and 3D-CNN \cite{funke2019using}, have been applied to endoscopic vision data. RAS datasets consist of synchronized data streams. The incorporation of multiple types of data, including robot kinematics, endoscopic vision, and system events (e.g., camera follow: a binary variable indication of if the endoscope is moving), can improve surgical state estimation accuracy in methods such as Latent Convolutional Skip-Chain CRF \cite{lea2016learning} and Fusion-KVE \cite{qin2020temporal}. 

\begin{figure}
    \centering
    \includegraphics[width=8cm]{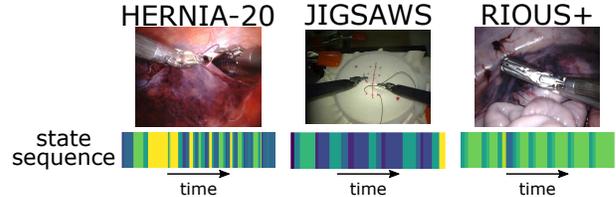}
    \vskip -0.07 true in
    \caption{{\bf Top row:} typical endoscopic images from HERNIA-20 (left), JIGSAWS (middle) and RIOUS+ (right) datasets. {\bf Bottom row:} related surgical state sequence samples, where each color represents a different surgical state.}
    \label{fig:Figure1}
    \vskip -0.25 true in
\end{figure}

Prior surgical state estimators relied heavily on RAS datasets for model fitting/training. Limitations in the dataset can be propagated (and perhaps amplified) to the estimator, possibly resulting in a lack of robustness and cross-domain generalizability \cite{lea2016learning}. Many surgical activity datasets are derived from highly uniform tasks performed using the same technique in only one setting. E.g., the JHU-ISI Gesture and Skill Assessment Working Set (JIGSAWS) \cite{gao2014jhu} suturing task was obtained in a bench-top setting, with suturing performed on marked pads (Fig. \ref{fig:Figure1}). Valuable anatomical background visual information is not present in the training data, which may lead to errors when the estimator is applied in real-world surgeries. Moreover, state estimators that are trained on datasets devoid of endoscope motion do not generalize well to new endoscopic views. Endoscope movements are frequent and spontaneous in real-world RAS. Additionally, operators in existing surgical activity datasets typically perform the task with the same technique, or were instructed to follow a predetermined workflow, which limits variability among trials. These limitations can cause state estimators to overfit to the techniques presented during training, and make inaccurate associations between surgical states and specific placements of surgical instruments and/or visual layout, instead of truely relevant features.

In real-world RAS tasks, endoscope lighting and viewing angles, surgical backgrounds, and patient health condition vary considerably among trials, as do state transition probabilities. We consider these variations as potential {\em nuisance factors} that increase the training difficulty of a robust surgical state estimator. Moreover, surgeons may employ diverse techniques to perform the same surgical task depending on patient condition and surgeon preferences. While the effects of nuisances and technique variations on estimation accuracy can be reduced by a large and diverse real-world RAS dataset, such datasets are costly to acquire.

The combination of limited data and high diversity calls for more robust state estimation training methods, as state-of-the-art methods are not accurate enough for adoption in the safety critical field of RAS. Surgical state estimation can be made invariant to irrelevant nuisances and surgeon techniques if latent representations of the input data contain minimal information about those factors \cite{xie2017controllable}. {\em Invariant representation learning} (IRL) has been an active research topic in computer vision, where robustness is achieved through {\em invariance induction} \cite{xie2017controllable, zemel2013learning, achille2018emergence, jaiswal2018unsupervised, jaiswal2019unified}. Zemel et al. proposed a supervised adversarial model to achieve fair classification under the two competing goals of encoding the input data correctly and obfuscating the group to which the data belongs \cite{zemel2013learning}. A regularized loss function using information bottleneck also induces invariance to nuisance factors \cite{achille2018emergence}. Jaiswal et al. described an adversarial invariance framework in which nuisance factors are distinguished through disentanglement \cite{jaiswal2018unsupervised}, and bias is distinguished through the competition between goal prediction and bias obfuscation \cite{jaiswal2019unified}. Previous work on IRL via adversarial invariance in time series data focused mostly on speech recognition \cite{hsu2019niesr, peri2020robust}. RAS data, arising from multiple sources, provides a new domain for IRL of high-dimensional noisy time series data.

\textbf{Contributions}: We propose {\em StiseNet}, a surgical state estimation model that is largely invariant to nuisances and variations in surgical techniques. StiseNet's adversarial design pits two composite models against each other to yield an invariant latent representation of the endoscopic vision, robot kinematics, and system event data. StiseNet learns a split representation of the input data through the competitive training of state estimation and input data reconstruction, and the disentanglement between essential information and nuisance. The influence of surgeon technique is excluded by adversarial training between state estimation and the obfuscation of a latent variable representing the technique type. StiseNet training does not require any additional annotation apart from surgical states. Our main contributions include:

\begin{itemize}
    \item{} An adversarial model design that promotes invariance to nuisance and surgical technique factors in RAS data.  
    \item{} A process to learn invariant latent representations of real-world RAS data streams, minimizing the effect of factors such as patient condition and surgeon technique.
    \item Improving frame-wise surgical state estimation accuracy for online and offline real-world RAS tasks by up to 7\%, which translates to a 28\% relative error reduction.
    \item Combining semantic segmentation with endoscopic vision to leverage a richer visual feature representation.
    \item{} Demonstrating the method on 3 RAS datasets.
\end{itemize}

StiseNet is evaluated and demonstrated using JIGSAWS suturing \cite{gao2014jhu}, RIOUS+ \cite{qin2020temporal}, and a newly collected HERNIA-20 dataset containing real-world hernia repair surgeries. StiseNet outperforms state-of-the-art surgical state estimation methods and improves frame-wise state estimation accuracy to 84\%. This level of error reduction is crucial for state estimation to gain adoption in RAS. StiseNet also accurately recognizes actions in a real-world RAS task even when a specific technique was not present in the training data. 

\section{Methods}

\begin{table}[b]
    \vskip -0.2 true in
\begin{tabular}{c|c}
Notation       & Description                                                                                                                                                                       \\ \hline
$\pmb{H}$      & \begin{tabular}[c]{@{}c@{}}Concatenated vision, kinematics, and event features\\ $\pmb{H} = \{\pmb{h}^{vis}, \pmb{h}^{kin}, \pmb{h}^{evt}\}$\end{tabular}                         \\
$s$            & Surgical state                                                                                                                                                                    \\
$T_{obs}$    & Observational window size                                                                                                                                    \\
$\pmb{e}_1$    & All factors pertinent to the estimation of $s$                                                                                                                                    \\
$\pmb{e}_2$    & \begin{tabular}[c]{@{}c@{}}All other factors (nuisance factors), which are of \\ no interest to goal variable estimation \cite{basu2011elimination}\end{tabular} \\
$l$            & Latent variable (type of surgical technique)                                                                                                                                      \\
$\overline{d}$ & Mean silhouette coefficient quantifying clustering quality                                                                                                                        \\ \hline
$\emph{E}$     & Encoder encodes $\pmb{H}$ into $\pmb{e}_1$ and $\pmb{e}_2$                                                                                                                        \\
$\emph{M}$     & Estimator infers $s$ from $\pmb{e}_1$                                                                                                                                             \\
$\psi$         & Dropout                                                                                                                                                                           \\
$\emph{R}$     & Reconstructer attempts to reconstruct $\pmb{H}$ from $[\psi(\pmb{e}_1), \pmb{e}_2]$                                                                                               \\
$\emph{f}_1$   & Disentangler infers $\pmb{e}_2$ from $\pmb{e}_1$                                                                                                                                  \\
$\emph{f}_2$   & Disentangler infers $\pmb{e}_1$ from $\pmb{e}_2$                                                                                                                                  \\
$\emph{D}$     & Discriminator estimates $l$ from $\pmb{e}_1$                                                                                                                                     
\end{tabular}
\caption{Key variables, concepts, and notation.}
\vskip -0.15 true in
\end{table}

StiseNet (Figs. \ref{fig:Figure2} and \ref{fig:Figure3}) accepts synchronized data streams of endoscopic vision, robot kinematics, and system events as inputs. To efficiently learn invariant latent representations of noisy data streams, we adopt an adversarial model design loosely following Jaiswal et al. \cite{jaiswal2019unified} but with model architectures more suitable for time series data. Jaiswal et al.'s adversarial invariance framework for image classification separates useful information and nuisance factors, such as lighting conditions, before performing classification. StiseNet extends this idea by separating learned features from RAS time series data into desired information for state estimation ($\pmb{e}_1$) and other information ($\pmb{e}_2$). Estimation is performed using $\pmb{e}_1$ to eliminate the negative effects of nuisances and variations in surgical techniques. LSTM computational blocks are used for feature extraction and surgical state estimation. LSTMs learn memory cell parameters that govern when to forget/read/write the cell state and memory \cite{dipietro2016recognizing}. They therefore better capture temporal correlations in time series data.  StiseNet's components and training procedure are described next. Table I lists key concepts and notation.

\subsection{Feature extraction}

\begin{figure}
    \centering
    \medskip
    \includegraphics[width=8.62cm]{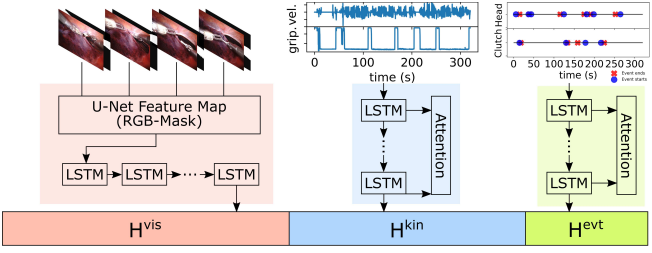}
    \caption{Features $\pmb{h}^{vis}$, $\pmb{h}^{kin}$, and $\pmb{h}^{evt}$ are respectively extracted from endoscopic vision, robot kinematics, and system events. A semantic mask is appended to the endoscopic vision data to form an RGB-Mask vision input.}
    \label{fig:Figure2}
    \vskip -0.28 true in
\end{figure}

Fig. \ref{fig:Figure2} depicts the extraction of features from endoscopic vision, robot kinematics, and system events data. Visual features are extracted by a CNN-LSTM model \cite{qin2020davincinet,yu2018learning}. To eliminate environmental distractions in the endoscopic view, a previously trained and frozen surgical scene segmentation model based on U-Net \cite{ronneberger2015u} extracts a pixel-level semantic mask for each frame. We use two scene classes: tissue and surgical instrument. The semantic mask is concatenated to the unmodified endoscope image as a fourth image channel. This RGB-Mask image $\pmb{I}_t \in \mathbb{R}^{h \times w \times 4}$ is then input to the CNN-LSTM. We implemented a U-Net-style feature map to extract visual features, $\pmb{x}_t^{vis}$, since a condensed surgical scene representation can be taken advantage of by adapting U-Net weights of the semantic segmentation model trained on a large endoscopic image dataset. We implemented an LSTM encoder to better capture temporal correlations in visual CNN features. This helps the visual processing system to extract visual features that evolve in time. At time $t$, a visual latent state, $\pmb{h}_t^{vis} \in \mathbb{R}^{n_{vis}}$, is extracted with the LSTM model.

Kinematics data are recorded from the Universal Patient-Side Manipulator (USM) of the da Vinci\textsuperscript{\textregistered} Surgical System. Kinematics features are extracted using an LSTM encoder with attention mechanism \cite{qin2017dual} to identify the important kinematics data types \cite{qin2020davincinet}. A multiplier $\pmb{\alpha}_t$, whose elements weight each type of kinematics data, was learned as follows:
\begin{equation}
    \pmb{\alpha}_t=softmax \left\{\pmb{u}^T tanh\left(\pmb{W}(\pmb{h}^{kin}_{t-1}, \pmb{c}^{kin}_{t-1}) + \pmb{V}\pmb{X}_t^{kin}\right)\right\},
\end{equation} 
where $\pmb{h}_{t-1}^{kin}$ is the latent state from the previous frame,  $\pmb{c}_{t-1}^{kin}$ is the LSTM cell state, and $\pmb{X}_t^{kin} = (\pmb{x}_{t-T_{obs}+1}^{kin}, ..., \pmb{x}_t^{kin})$ denote the kinematic data inputs. $\pmb{u}$, $\pmb{W}$, and $\pmb{V}$ are learnable parameters. The weighted kinematics data feature vector $\pmb{h}_t^{kin} \in \mathbb{R}^{n_{kin}}$ is calculated as:
\begin{equation}
    \pmb{h}_t^{kin} = LSTM(\pmb{h}_{t-1}^{kin}, \pmb{\alpha}_t \cdot \pmb{x}_t^{kin})
\end{equation}
The da Vinci\textsuperscript{\textregistered} Xi Surgical System also provides system event data (details in Section \ref{experimental_evaluation}). The event features $\pmb{h}_t^{evt}$ are extracted via the same method as kinematics.

\subsection{Feature encoder and Surgical state estimator}
\begin{figure*}
    \centering
    \medskip
    \includegraphics[width=13cm]{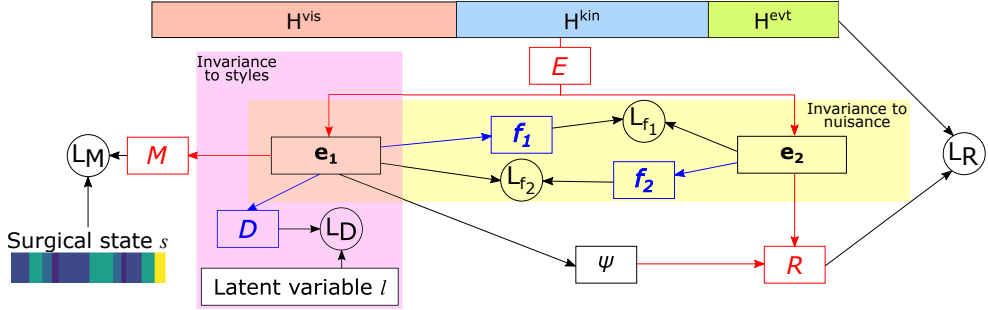}
    \caption{StiseNet training architecture. Symbols for the estimator components $P1=\{\emph{E}, \emph{M}, \emph{R}\}$ are \textcolor{red}{red}, the adversarial component $P2=\{\emph{f}_1, \emph{f}_2, \emph{D}\}$ is \textcolor{blue}{blue}, and training loss calculations are \textcolor{black}{black}. $P2$ implements invariance to nuisance (yellow shading) and surgical techniques (pink shading). RAS data features $\pmb{H}$ are divided into information essential for state estimation, $\pmb{e}_1$, and other information $\pmb{e}_2$. $\pmb{H}$ is reconstructed from $\psi(\pmb{e}_1)$ and $\pmb{e}_2$, where $\psi$ is dropout.}
    \label{fig:Figure3}
    \vskip -0.2 true in
\end{figure*}
As shown in Fig. \ref{fig:Figure3}, Encoder $\emph{E}$ extracts useful information for estimation from the latent feature data $\pmb{H}$. If we assume that $\pmb{H}$ is composed of a set of factors of variation, then $\pmb{H}$ it is composed of mutually exclusive subsets: 
\begin{itemize}
    \item $\pmb{e}_1$: all the factors pertinent to the estimation of the goal variable (the current surgical state $s$);
    \item $\pmb{e}_2$: all other factors (nuisance factors), which are of no interest to goal variable estimation \cite{basu2011elimination}. 
\end{itemize}
Encoder $\emph{E}$ is a function trained to partition $\pmb{H}$: $[\pmb{e}_1,\pmb{e}_2] = \emph{E}(\pmb{H})$. A fully-connected (FC) layer maps $\pmb{H}$ to $\pmb{e}_1$, and another FC layer maps $\pmb{H}$ to $\pmb{e}_2$. Once distinguished, the surgical state $s$ at time $t$ is estimated from the history of the useful signal $\{\pmb{e}_{1,t-T_{obs}+1}, \ldots, \pmb{e}_{1,t}\}$ using an LSTM decoder $\pmb{M}$ following \cite{qin2020davincinet}. By learning the parameters in $M$ using $\pmb{e}_1$ instead of $\pmb{H}$, we avoid learning inaccurate associations between nuisance factors and the goal variable.

\subsection{Learning an invariant representation}

The invariance induction to nuisance and technique factors is learned via \emph{competition} and \emph{adverseness} between model components \cite{goodfellow2014generative} (yellow and pink shaded components in Fig. \ref{fig:Figure3}). While \emph{M} encourages the pooling of  factors relevant to surgical state estimation in signal $\pmb{e}_1$, a reconstructor \emph{R} (a function implemented as an FC layer) attempts to reconstruct from the separated signals. Dropout $\psi$ is added to $\pmb{e}_1$ to make it an unreliable source to reconstruct $\pmb{H}$ \cite{jaiswal2018unsupervised}. This configuration of signals prevents a convergence to the trivial solution where $\pmb{e}_1$ monopolizes all information, while $\pmb{e}_2$ contains none. The mutual exclusivity between $\pmb{e}_1$ and $\pmb{e}_2$ is achieved through adversarial training. Two FC layers $\emph{f}_1$ and $\emph{f}_2$ are implemented as \emph{disentanglers}. $\emph{f}_1$ attempts to infer $\pmb{e}_2$ from $\pmb{e}_1$, while $\emph{f}_2$ infers $\pmb{e}_1$ from $\pmb{e}_2$. To achieve mutual exclusivity, we should not be able to infer $\pmb{e}_1$ from $\pmb{e}_2$ or vise versa. Hence, the losses of $\emph{f}_1$ and $\emph{f}_2$ must be maximized. This leads to an adversarial training objective \cite{mathieu2016disentangling}. The loss function with invariance to nuisance factors is:
\begin{align}\label{eq:Equation4}
    L_{nuis} &= \alpha L_{M}\left(s,\emph{M}(\pmb{e}_1)\right) + \beta L_{R}\left(\pmb{H},\emph{R}(\pmb{e}_2, \psi(\pmb{e}_1))\right) \\&\quad \nonumber + \gamma\left( L_{f}\left(\pmb{e}_1,\emph{f}_1(\pmb{e}_2)\right) + L_{f}\left(\pmb{e}_2,\emph{f}_2(\pmb{e}_1)\right)\right) \nonumber
\end{align}
where $\alpha$, $\beta$, and $\gamma$ respectively weight the adversarial loss terms \cite{mathieu2016disentangling} associated with architectural components $M$, $R$, and disentanglers $f_1$ and $f_2$. The training objective with invariance to nuisance factors is a minimax game \cite{maschler2013game, goodfellow2014generative}:
\begin{equation}\label{eq:Equation5}
    \min_{P1} \, \max_{P2_{nuis}} \, L_{nuis}
\end{equation}
where the loss of component $P1 = \{\emph{E}, \emph{M},\emph{R}\}$ is minimized while the loss of $P2_{nuis} = \{\emph{f}_1,\emph{f}_2\}$ maximized.
 
Besides the presence of nuisance factors, variability in $\pmb{H}$ could also arise from variability in surgical techniques. Variations in technique may not be entirely separable by an invariance to nuisance factors, as they may be correlated to the surgical state. StiseNet therefore adopts an {\em adversarial debiasing} design \cite{zhang2018mitigating} that deploys a {\em discriminator} $\emph{D}:\pmb{e}_1 \rightarrow l$ for surgical technique invariance. The latent variable $l$ represents the type of technique employed to perform a surgical task. $l$ is a trial-level categorical attribute that is inferred by k-means clustering of kinematics time series training data based on a dynamic time warping distance metric (function $\phi$)\cite{sakoe1978dynamic}. The clusters represent different surgical techniques used in the training trials. The optimal number of clusters $k$ is dataset-specific. To determine it, we implemented the {\em elbow method} using inertia \cite{forgy1965cluster} and the {\em silhouette method} \cite{rousseeuw1987silhouettes}. The inertia is defined as the sum of squared distances between each cluster member and its cluster center\cite{forgy1965cluster} for all clusters. The inertia decreases as $k$ increases, and the elbow point is a relatively optimal $k$ value \cite{forgy1965cluster}. The silhouette coefficient $d_i$ for time series $i$ is:
\begin{equation}
    d_i = \frac{\underset{j}{mean}\left( \min\limits_{m \notin C_i} \sum\limits_{j \in C_m} \phi(i,j)
    - \sum\limits_{j \in C_i, j \neq i} \phi(i,j)\right)}
    {max(a_i,b_i)}
\end{equation}
where $C_i$ is the cluster of time series $i$. The operation $\min\limits_{m \notin C_i}$ represents the closest time series to $i$ that does not belong to $C_i$. We used the mean silhouette coefficient among all time series $\overline{d}$ to select $k$. $\overline{d}$ is a measure of how close each data point in one cluster is to data points in the nearest neighboring clusters. The $k$ with the highest $\overline{d}$ is the optimal number of clusters. The loss function with invariance to both nuisance and surgical techniques is then:
\begin{equation}\label{eq:Equation9}
    L  = L_{nuis} + \delta L_{D}\left(l,\emph{D}(\pmb{e}_1)\right)
\end{equation}
where $\delta$ is the weight associated with the discriminator loss. The term $P2$ contains an additional term: $\tilde{P2} = \{\emph{f}_1,\emph{f}_2, \emph{D}\}$:  
\begin{equation}\label{eq:Equation10}
    \min_{P1} \, \max_{\tilde{P2}} \, L.
 \end{equation}

\subsection{Training and inference}

StiseNet's feature extraction components were trained following \cite{qin2020temporal}. Specifically, the first three channels of the top layer in U-Net visual feature map were initialized with the weights from the surgical scene segmentation model. The visual input was resized to $h = 256$ and $w = 256$. The extracted features have dimensions $n_{vis} = 40$, $n_{kin} = 40$, and $n_{evt} = 4$, which were determined using grid search. All data sources are synchronized at 10Hz with $T_{obs} = 20\ samples = 2 sec$. The optimal cluster number, $k$, for JIGSAWS, RIOUS+, and HERNIA-20 were 9, 7, and 4, respectively. The temporal clustering process was repeated to ensure reproducibility due to the randomness in initialization. Section \ref{sec:results} described how $k$ is determined in these datasets.

StiseNet is trained end-to-end with the minimax objectives (Eq.s \ref{eq:Equation5} and \ref{eq:Equation10}). We used the categorical cross-entropy loss for $L_{M}$ and $L_{D}$. $L_{f}$ and $L_{R}$ are mean squared error loss. $\psi$ is a dropout \cite{srivastava2014dropout} with the rate of 0.4, 0.1, and 0.4 for JIGSAWS, RIOUS+, and HERNIA-20, respectively. To effectively train the adversarial model, we applied a scheduled adversarial optimizer \cite{goodfellow2014generative}, in which a training batch is passed to either $P1$ or $P2$ while the other component's weights are frozen. The alternating schedule was found by grid search to be 1:5. 

\section{Experimental Evaluation}\label{experimental_evaluation}

We evaluated StiseNet's performance on the JIGSAWS suturing \cite{gao2014jhu}, RIOUS+ \cite{qin2020temporal}, and a newly collected HERNIA-20 dataset, respectively. These datasets were annotated with manually determined lists of fine-grained states (Table II).

\subsection{Datasets}

\textbf{JIGSAWS}: The JIGSAWS \cite{gao2014jhu} bench-top suturing task includes 39 trials by eight surgeons partaking in nine surgical actions. We used the endoscopic vision and USM's kinematics (gripper angle,  translational and rotational positions and velocities) data. There was no system events data. The tooltips' orientation matrices were converted to Euler angles. 

\textbf{RIOUS+}: The RIOUS+ dataset, introduced in \cite{qin2020temporal, qin2020davincinet}, captures 40 trials of an ultrasound scanning task on a da Vinci Xi\textsuperscript{\textregistered} Surgical System by five users in a mixture of bench-top (27) and OR (13) trials. Eight states represent user actions or environmental changes. Endoscopic vision, USM kinematics, and six binary system events serve as inputs--see \cite{qin2020temporal}. A finite state machine model of the task was determined prior to data collection. The operators were instructed to strictly follow this predetermined task workflow and to ignore environmental disruptions. The action sequences and techniques are therefore highly structured and similar across trials. While it includes more realistic RAS elements, such as OR settings and endoscope movements, RIOUS+ lacks the behavioral variability of real-world RAS data.

\textbf{HERNIA-20}: The HERNIA-20 dataset contains 10 fully anonymized real-world robotic transabdominal preperitoneal inguinal hernia repair procedures performed by surgeons on da Vinci Xi \textsuperscript{\textregistered} Surgical Systems. For performance evaluation, we selected a running suturing task performed to re-approximate the peritoneum, which contains 11 states. The endoscopic vision, USM kinematics, and system events are used as inputs. Because HERNIA-20 captures real-world RAS performed on patients, the robustness of surgical state estimation models can be fully examined.

\begin{table}[]
\vskip -0.15 true in
\footnotesize
\centering
\begin{tabular}{ccc}
\hline
\multicolumn{1}{c|}{Gesture} & \multicolumn{1}{c|}{\textbf{JIGSAWS Suturing Dataset}}                              & Duration (s) \\ \hline
\multicolumn{1}{c|}{G1}        & \multicolumn{1}{c|}{Reaching for the needle with right hand}  & 2.2          \\
\multicolumn{1}{c|}{G2}        & \multicolumn{1}{c|}{Positioning the tip of the needle}        & 3.4          \\
\multicolumn{1}{c|}{G3}        & \multicolumn{1}{c|}{Pushing needle through the tissue}        & 9.0          \\
\multicolumn{1}{c|}{G4}        & \multicolumn{1}{c|}{Transferring needle from left to right}   & 4.5          \\
\multicolumn{1}{c|}{G5}        & \multicolumn{1}{c|}{Moving to center with needle in grip}     & 3.0          \\
\multicolumn{1}{c|}{G6}        & \multicolumn{1}{c|}{Pulling suture with left hand}            & 4.8          \\
\multicolumn{1}{c|}{G7}        & \multicolumn{1}{c|}{Orienting needle}                         & 7.7          \\
\multicolumn{1}{c|}{G8}        & \multicolumn{1}{c|}{Using right hand to help tighten suture}  & 3.1          \\
\multicolumn{1}{c|}{G9}        & \multicolumn{1}{c|}{Dropping suture and moving to end points} & 7.3          \\
\hline
\multicolumn{1}{c|}{State}  & \multicolumn{1}{c|}{\textbf{RIOUS+ Dataset}}                              & Duration (s) \\ \hline
\multicolumn{1}{c|}{S1}        & \multicolumn{1}{c|}{Probe released, out of endoscopic view}   & 6.3         \\
\multicolumn{1}{c|}{S2}        & \multicolumn{1}{c|}{Probe released, in endoscopic view}       & 7.6         \\
\multicolumn{1}{c|}{S3}        & \multicolumn{1}{c|}{Reaching for probe}                       & 3.1          \\
\multicolumn{1}{c|}{S4}        & \multicolumn{1}{c|}{Grasping probe}                           & 1.1          \\
\multicolumn{1}{c|}{S5}        & \multicolumn{1}{c|}{Lifting probe up}                         & 2.4          \\
\multicolumn{1}{c|}{S6}        & \multicolumn{1}{c|}{Carrying probe to tissue surface}         & 2.3          \\
\multicolumn{1}{c|}{S7}        & \multicolumn{1}{c|}{Sweeping}                                 & 5.1          \\
\multicolumn{1}{c|}{S8}        & \multicolumn{1}{c|}{Releasing probe}                          & 1.7          \\         
\hline
\multicolumn{1}{c|}{State} & \multicolumn{1}{c|}{\textbf{HERNIA-20 Dataset}}                            & \multicolumn{1}{l}{Duration (s)} \\ \hline
\multicolumn{1}{c|}{S1}       & \multicolumn{1}{c|}{Reaching for the needle}                & 3.9                              \\
\multicolumn{1}{c|}{S2}       & \multicolumn{1}{c|}{Positioning the tip of the needle}      & 3.3                              \\
\multicolumn{1}{c|}{S3}       & \multicolumn{1}{c|}{Pushing needle through the tissue}      & 4.2                              \\
\multicolumn{1}{c|}{S4}       & \multicolumn{1}{c|}{Pulling tissue with left hand}          & 3.6                              \\
\multicolumn{1}{c|}{S5}       & \multicolumn{1}{c|}{Transferring needle from left to right} & 3.7                              \\
\multicolumn{1}{c|}{S6}       & \multicolumn{1}{c|}{Orienting needle}                       & 6.6                              \\
\multicolumn{1}{c|}{S7}       & \multicolumn{1}{c|}{Pulling suture with left hand}          & 5.8                              \\
\multicolumn{1}{c|}{S8}       & \multicolumn{1}{c|}{Pulling suture with right hand}         & 4.8                              \\
\multicolumn{1}{c|}{S9}       & \multicolumn{1}{c|}{Transferring needle from right to left} & 4.6                              \\
\multicolumn{1}{c|}{S10}      & \multicolumn{1}{c|}{Using right hand to tighten suture}     & 4.3                              \\
\multicolumn{1}{c|}{S11}      & \multicolumn{1}{c|}{Adjusting endoscope}                    & 3.8                             
\end{tabular}
\caption{Datasets State Descriptions and Mean Duration}
\vskip -0.25 true in
\end{table}

\subsection{Metrics}

The quality of the learned invariant representations of surgical states $\pmb{e}_1$ and other information $\pmb{e}_2$ is visually examined. Arrays of $\pmb{e}_1$ and $\pmb{e}_2$ in each state instance (a consecutive block of time frames of the same surgical state) are embedded in 2D space using the Uniform Manifold Approximation and Projection (UMAP) algorithm \cite{mcinnes2018umap} - a widely-adopted dimension reduction and visualization method that preserves more of the global structure of the data.

We used the percentage of accurately identified frames in a test set to evaluate each model's surgical state estimation accuracy. Model performance was evaluated in non-causal and causal settings. In a non-causal setting, the model can use information from future time frames, which is suitable for post-operative analyses. In causal settings, the model only has access to the current and preceding time frames. Surgical state estimation is harder in the causal setting; however, it is a more useful evaluation metric for real-time applications.

We used the source code provided by the authors of the comparison methods when the model performance of a particular setting or dataset was not available\cite{lea2016temporal,dipietro2016recognizing} and performed training and evaluation ourselves. JIGSAWS suturing and RIOUS+ datasets were evaluated using \emph{Leave One User Out} (LOUO) \cite{gao2014jhu}, while HERNIA-20 was evaluated using 5-fold cross validation, since each trial's surgeon ID is not available due to privacy protection. 

\subsection{Ablation Study}

We compared StiseNet against its two ablated versions: StiseNet-Non Adversarial (StiseNet-NA) and StiseNet-Nuisance Only (StiseNet-NO). StiseNet-NA omits the adversarial component P2 entirely (the yellow and pink-shaded areas in Fig. \ref{fig:Figure3}) and uses $\pmb{H}$ for estimation with Estimator $M: \pmb{H}_t \rightarrow s_t $. StiseNet-NO separates useful information and nuisance factors, but excludes the invariance to surgical techniques (pink-shaded area in Fig. \ref{fig:Figure3}). The ablation study demonstrates the necessity of the adversarial model design and individual contributions of each model component towards a more accurate surgical state estimation.

\section{Results and Discussions} \label{sec:results}

\begin{figure*}
    \centering
    \medskip
    \includegraphics[width=16.5cm]{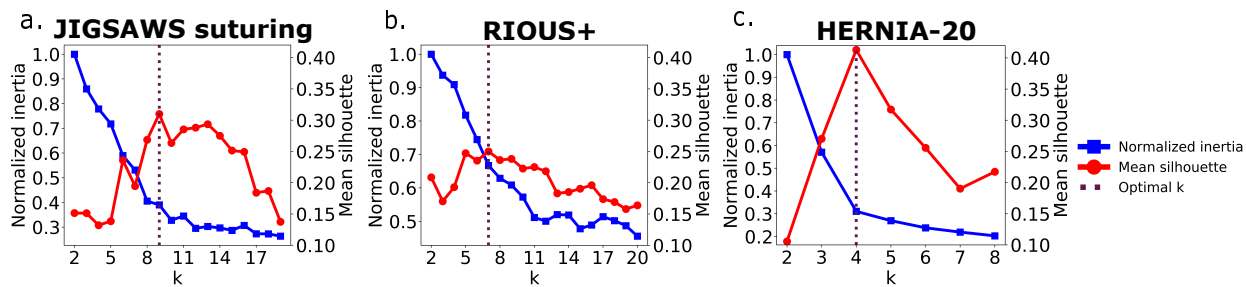}
    \caption{Normalized inertia (with respect to the maximum value) and mean silhouette coefficient as functions of the number of clusters $k$ for each dataset. The vertical dotted line indicates the optimal $k$ (the maximum mean silhouette coefficient).}
    \label{fig:Figure4}
    \vskip -0.2 true in
\end{figure*}
\begin{table}[b]
\vskip -0.15 true in
\setlength\tabcolsep{4pt}
\begin{tabular}{ccccc}
\multicolumn{5}{c}{\textbf{Non-causal}}                                                                                                                   \\
\multicolumn{1}{c|}{}            & \multicolumn{1}{c|}{Input data} & \multicolumn{1}{c|}{JIGSAWS} & \multicolumn{1}{c|}{RIOUS+}        & HERNIA-20        \\ \hline
\multicolumn{1}{c|}{TCN\cite{lea2016temporal}}         & \multicolumn{1}{c|}{kin}             & \multicolumn{1}{c|}{79.6}             & \multicolumn{1}{c|}{82.0}          & 72.1          \\
\multicolumn{1}{c|}{TCN\cite{lea2016temporal}}         & \multicolumn{1}{c|}{vis}             & \multicolumn{1}{c|}{81.4}             & \multicolumn{1}{c|}{62.7}          & 61.5          \\
\multicolumn{1}{c|}{Bidir. LSTM\cite{dipietro2016recognizing}} & \multicolumn{1}{c|}{kin}             & \multicolumn{1}{c|}{83.3}             & \multicolumn{1}{c|}{80.3}          & 73.8          \\
\multicolumn{1}{c|}{LC-SC-CRF\cite{lea2016learning}}   & \multicolumn{1}{c|}{vis+kin}         & \multicolumn{1}{c|}{83.5}             & \multicolumn{1}{c|}{-}             & -             \\
\multicolumn{1}{c|}{3D-CNN\cite{funke2019using}}      & \multicolumn{1}{c|}{vis}             & \multicolumn{1}{c|}{84.3}             & \multicolumn{1}{c|}{-}             & -             \\
\multicolumn{1}{c|}{Fusion-KVE\cite{qin2020temporal}}  & \multicolumn{1}{c|}{vis+kin+evt}     & \multicolumn{1}{c|}{86.3}             & \multicolumn{1}{c|}{\textbf{93.8}} & 78.0          \\
\multicolumn{1}{c|}{StiseNet-NA} & \multicolumn{1}{c|}{vis+kin+evt}     & \multicolumn{1}{c|}{86.5}             & \multicolumn{1}{c|}{93.1}          & 80.0          \\
\multicolumn{1}{c|}{StiseNet-NO} & \multicolumn{1}{c|}{vis+kin+evt}     & \multicolumn{1}{c|}{87.9}             & \multicolumn{1}{c|}{90.3}          & 83.2          \\
\multicolumn{1}{c|}{StiseNet}    & \multicolumn{1}{c|}{vis+kin+evt}     & \multicolumn{1}{c|}{\textbf{90.2}}    & \multicolumn{1}{c|}{92.5}          & \textbf{84.1}
\end{tabular}
\caption*{Table III: State estimation performance in non-causal setting. JIGSAWS results did not include system events.}

\medskip

\setlength\tabcolsep{4pt}
\begin{tabular}{ccccc}
\multicolumn{5}{c}{\textbf{Causal}}                                                                                                                            \\
\multicolumn{1}{c|}{}             & \multicolumn{1}{c|}{Input data}  & \multicolumn{1}{c|}{JIGSAWS}       & \multicolumn{1}{c|}{RIOUS+}        & HERNIA-20     \\ \hline
\multicolumn{1}{c|}{TCN\cite{lea2016temporal}}          & \multicolumn{1}{c|}{vis}         & \multicolumn{1}{c|}{76.8}          & \multicolumn{1}{c|}{54.8}          & 58.3          \\
\multicolumn{1}{c|}{TCN\cite{lea2016temporal}}          & \multicolumn{1}{c|}{kin}         & \multicolumn{1}{c|}{72.4}          & \multicolumn{1}{c|}{78.4}          & 68.1          \\
\multicolumn{1}{c|}{Forward LSTM\cite{dipietro2016recognizing}} & \multicolumn{1}{c|}{kin}         & \multicolumn{1}{c|}{80.5}          & \multicolumn{1}{c|}{72.2}          & 69.8          \\
\multicolumn{1}{c|}{3D-CNN\cite{funke2019using}}       & \multicolumn{1}{c|}{vis}         & \multicolumn{1}{c|}{81.8}          & \multicolumn{1}{c|}{-}             & -             \\
\multicolumn{1}{c|}{Fusion-KVE\cite{qin2020temporal}}   & \multicolumn{1}{c|}{vis+kin+evt} & \multicolumn{1}{c|}{82.7}          & \multicolumn{1}{c|}{89.4} & 75.7          \\
\multicolumn{1}{c|}{StiseNet-NA}  & \multicolumn{1}{c|}{vis+kin+evt} & \multicolumn{1}{c|}{83.4}          & \multicolumn{1}{c|}{88.9}          & 77.3          \\
\multicolumn{1}{c|}{StiseNet-NO}  & \multicolumn{1}{c|}{vis+kin+evt} & \multicolumn{1}{c|}{84.1}          & \multicolumn{1}{c|}{88.9}          & 81.0          \\
\multicolumn{1}{c|}{StiseNet}     & \multicolumn{1}{c|}{vis+kin+evt} & \multicolumn{1}{c|}{\textbf{85.6}} & \multicolumn{1}{c|}{\textbf{89.5}}          & \textbf{82.7}
\end{tabular}
\caption*{Table IV: State estimation performance in a causal setting. JIGSAWS results did not include system events.}
\vskip -0.1 true in
\end{table}
\begin{figure*}
    \centering
    \medskip
    \includegraphics[width=17cm]{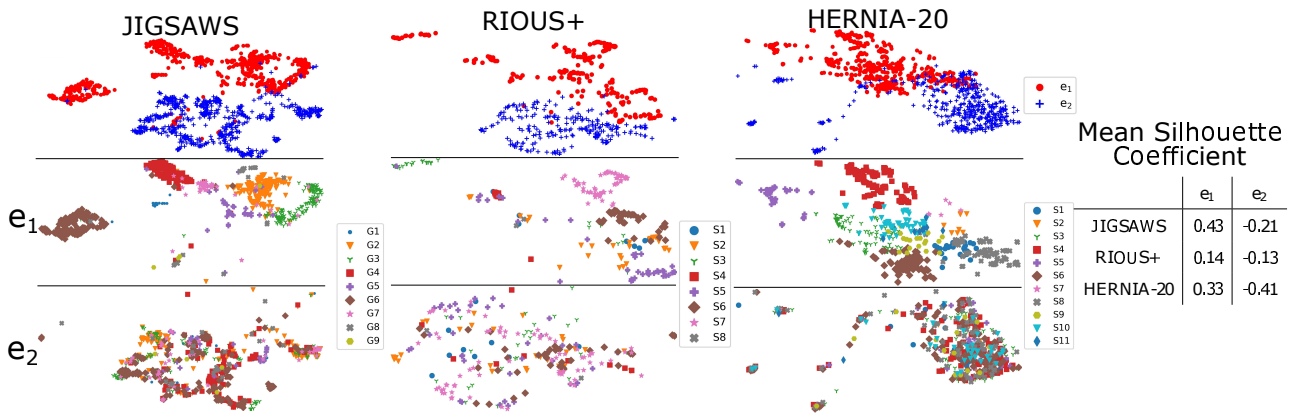}
    \caption{2D UMAP plots of information enclosed in $\pmb{e}_1$ and $\pmb{e}_2$ at each state instance. {\bf Top row:} $\pmb{e}_1$ and $\pmb{e}_2$ segregates into distinguishable clusters, which indicates little overlap in information. {\bf Middle row:} Information in $\pmb{e}_1$ color-coded by surgical states clusters relatively neatly. {\bf Bottom row:} Information in $\pmb{e}_2$ is more intertwined and non-distinguishable by state. The mean silhouette coefficient $\overline{d}$ of each graph is shown, with a larger $\overline{d}$ indicating better clustering quality.}
    \label{fig:umap}
    \vskip -0.2 true in
\end{figure*} 

\begin{figure*}
    \centering
    \medskip
    \includegraphics[width=15.5cm]{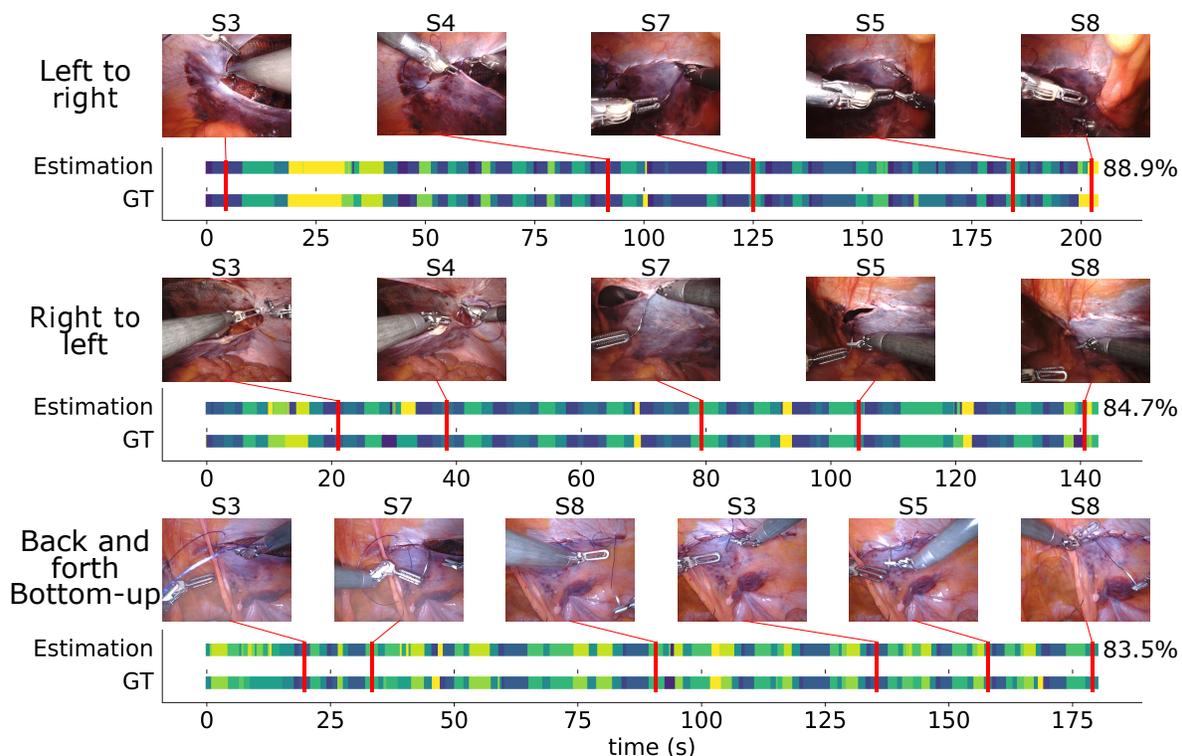}
    \vskip -0.1 true in
    \caption{Three HERNIA-20 trials from three technique clusters, and StiseNet's performances compared to ground truth (GT). Instances of the same state in different trials are substantially and visibly different; however, StiseNet correctly estimates them. Variations across trials arise from both nuisances and techniques. Potential sources of nuisances include but are not limited to lighting conditions, presence of fat or blood, peritoneum color, endoscope movements, etc.}
    \label{fig:Figure5}
    \vskip -0.25 true in
\end{figure*}

\begin{figure*}
    \centering
    \medskip
    \includegraphics[width=16cm]{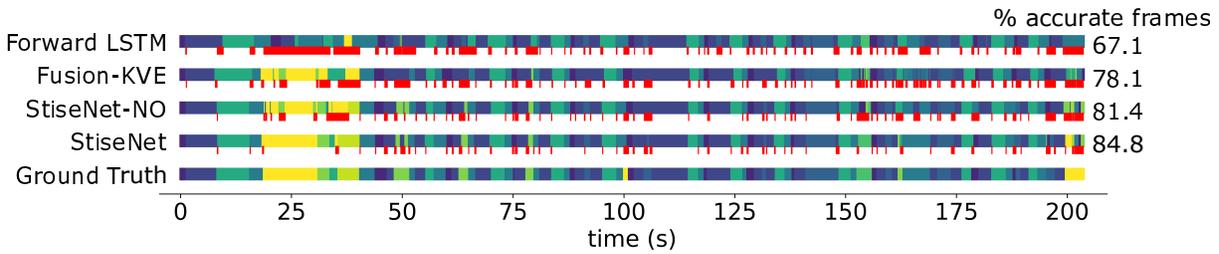}
    \caption{Example HERNIA-20 surgical state estimation results by forward LSTM \cite{dipietro2016recognizing}, Fusion-KVE \cite{qin2020temporal}, StiseNet-NO, and StiseNet, compared to ground truth. State estimation results (top) and discrepancies with ground truth in red (bottom) are shown in each block bar.}
    \label{fig:Figure6}
    \vskip -0.3 true in
\end{figure*}

Fig. \ref{fig:Figure4} plots for each dataset the {\em total inertia} and the {\em mean silhouette coefficient} $\overline{d}$ as functions of the number of clusters $k$. Fig. \ref{fig:umap} shows the UMAP visualizations of $\pmb{e}_1$ and $\pmb{e}_2$ for all surgical states. We compare both the non-causal (Table III) and causal (Table IV) performance of StiseNet with its ablated versions and prior methods. Fig. \ref{fig:Figure5} shows the variability in HERNIA-20 data through sample sequences from three technique clusters, each performed in a distinctively different style with environmental variances. Invariance of StiseNet to nuisances and surgical techniques is shown by its accurate surgical state estimations in the presence of visibly diverse input data. Fig. \ref{fig:Figure6} shows a sample state sequence from HERNIA-20 and the causal state estimation results using multiple methods, including forward LSTM \cite{dipietro2016recognizing}, Fusion-KVE \cite{qin2020temporal}, and the ablated and full versions of StiseNet.

As mentioned in Section II-C, the optimal number of clusters $k$ can be estimated from the elbow point of the inertia-$k$ curve, or the $k$ associated with the maximum mean silhouette coefficient $\overline{d}$. We implemented both methods and illustrate our choices of $k$ in Fig. \ref{fig:Figure4}. The optimal $k$ is easily identifiable for JIGSAWS and HERNIA-20 (Fig. \ref{fig:Figure4}a and \ref{fig:Figure4}c), with the largest $\overline{d}$ occurs near the "elbow" of the inertia-$k$ curve. A peak in the RIOUS+ mean silhouette coefficient curve is less evident (Fig. \ref{fig:Figure4}b). The optimal number of clusters need not match the number of operators, as the inter-personal characteristics are not the only accountable factor for the variations among trials. Intra-personal variations can affect clustering. E.g., JIGSAWS contains metadata corresponding to expert ratings of each trial \cite{gao2014jhu}: the ratings fluctuate among trials performed by the same surgeon. The optimal $k$ determined by kinematics data is somewhat robust against patient anatomy; however, a highly unique patient anatomy can lead surgeons to modify their maneuvers significantly. Such a trial could fall into a different technique cluster.

Fig. \ref{fig:umap} visualizes the 2D projections of $\pmb{e}_1$ and $\pmb{e}_2$. The first row shows that $\pmb{e}_1$ and $\pmb{e}_2$ separate neatly into two clusters for all datasets, validating the effectiveness of disentanglers $f_1$ and $f_2$ since $\pmb{e}_1$ and $\pmb{e}_2$ contain little overlapping information. Since $\pmb{e}_1$ contains useful information for state estimation, while $\pmb{e}_2$ does not, $\pmb{e}_1$ should be better segregated into clusters associated to each state. The second and third rows of Fig. \ref{fig:umap} (color-coded by surgical state) show cleanly segregated clusters for $\pmb{e}_1$, while the $\pmb{e}_2$ projections are not distinguishable by state. The {\em mean silhouette coefficient} for each graph also supports this observation. This strongly suggests that each surgical state has a unique representation in $\pmb{e}_1$, while $\pmb{e}_2$ contains little information useful for state estimation.

Table III and IV show non-causal and causal surgical state estimation performance of recently proposed methods and StiseNet (and its ablated versions). Both StiseNet and StiseNet-NO yield an improvement in frame-wise surgical state estimation accuracy for JIGSAWS suturing (up to 3.9\%) and HERNIA-20 (up to 7\%) under both settings, which shows the necessity and effectiveness of the adversarial model design. The non-causal performance of StiseNet on RIOUS+ is slightly worse compared to our Fusion-KVE method \cite{qin2020temporal}, which does not dissociate nuisance or style variables. This result can be explained by StiseNet's model design and training scheme. The added robustness of StiseNet against variations in background, surgical techniques, etc. comes at the cost of the increased training complexity associated with adversarial loss functions and minimax training. Surgeon techniques and styles vary in JIGSAWS, and more significantly in HERNIA-20. Nuisance factors (tissue deformations, endoscopic lighting conditions and viewing angles, etc.) also vary considerably among trials and users in HERNIA-20. However, since RIOUS+ users were instructed to strictly follow a predetermined workflow, there are few nuisance and technique factors. The disentanglement between essential information $\pmb{e}_1$ and other information $\pmb{e}_2$ is therefore less effective. This hypothesis is supported by the observation that the dropout rate required for StiseNet training covergence is 0.1 for RIOUS+, whereas JIGSAWS and HERNIA-20 training converged with a dropout rate of 0.4. A lower dropout rate indicates that $\pmb{e}_2$ contains little information despite the dropout's effort to avoid the trivial solution. Additionally, the uniformity across RIOUS+ participants results in a nearly constant mean silhouette coefficient (Fig. \ref{fig:Figure4}b). Hence, StiseNet's invariance properties cannot be fully harnessed, explaining its less competitive performance in RIOUS+ as compared to the real-world RAS data of HERNIA-20.
%

In real-world RAS, surgeons may use different techniques to accomplish the same task. Fig. \ref{fig:Figure5} shows three HERNIA-20 trials with distinctive suturing geometries: suturing from left to right, from right to left, and back and forth along a vertical seam. These trials fall into three clusters. We show images from instances of states S3, S4, S5, S7 and S8 in each trial. These images of different instances of the same state vary greatly not just in technique and instrument layout, but also in nuisance factors such as brightness and endoscope angles. Yet, StiseNet accurately estimates the surgical states due to its invariant latent representation of the input data.

Fig. \ref{fig:Figure6} demonstrates StiseNet's robustness during rapid and unpredictable state transitions in a real-world RAS suturing task. We compare the causal estimation performance of Forward-LSTM, Fusion-KVE, the ablated, and full versions of StiseNet against ground truth. Forward-LSTM, which only uses kinematics data, has a block of errors from 20s to 30s since it cannot recognize the "adjusting endoscope" state due to a lack of visual and event inputs. When those inputs are added, Fusion-KVE and StiseNet recognize this state. Fusion-KVE still shows a greater error rate due to limited training data with high environmental diversity, which reflects Fusion-KVE's vulnerability to nuisance and various surgical techniques. StiseNet-NO shows fewer error blocks: yet it is still affected by different technique types. The higher estimation accuracy of StiseNet shows its technique-agnostic robustness in real-world RAS, even with a small training dataset that contains behavioral and environmental diversity.

\section{Conclusions and Future Work}

This paper focused on improving the accuracy of surgical state estimation in real-world RAS tasks learned from limited amounts of data with high behavioral and environmental diversity. We proposed \emph{StiseNet}: an adversarial learning model with an invariant latent representation of RAS data. StiseNet was evaluated on three datasets, including a real-world RAS dataset that includes different surgical techniques carried out in highly diverse environments. StiseNet improves the state-of-the-art performance by up to 7\%. The improvement is significant for the real-world running suture tasks, which benefit greatly from invariance to surgical techniques, environments, and patient anatomy. Ablation studies showed the effectiveness of the adversarial model design and the necessity of invariance inductions to both nuisance and technique factors. StiseNet training does not require additional annotation apart from the surgical states. We plan to further investigate alternative labelling methods of surgical techniques and the invariance induction to other latent variables such as surgeon ID, surgeon levels of expertise, etc. Due to the limited data availability, StiseNet has only been evaluated on small datasets. Adding more trials to HERNIA-20 will allow us to evaluate StiseNet more comprehensively. To further improve estimation accuracy, StiseNet's neural network architectures may be further optimized for a better learning of temporal correlations within data. We also plan to incorporate longer-term context information \cite{ban2020aggregating, zhang2020symmetric}. StiseNet's accurate and robust surgical state estimation could also aide the development of surgeon-assisting functionalities and shared control systems in RAS. 




{\bf ACKNOWLEDGMENT:}
This work was funded by Intuitive Surgical, Inc. We would like to thank Dr. Seyedshams Feyzabadi, Dr. Azad Shademan, Dr. Sandra Park, Dr. Humphrey Chow, and Dr. Wenqing Sun for their support.

\bibliographystyle{IEEEtran}
\bibliography{IEEEabrv,IEEEexample}

\end{document}